\documentclass[conference]{IEEEtran}
\IEEEoverridecommandlockouts
\usepackage{cite}
\usepackage{amsmath,amssymb,amsfonts}
\usepackage{comment}
\usepackage{amsfonts}
\usepackage{algorithmic}

\usepackage[dvipdfmx]{graphicx}
\usepackage{textcomp}
\usepackage{xcolor}
\usepackage{subfigure}
\usepackage{amsthm} 

\def\BibTeX{{\rm B\kern-.05em{\sc i\kern-.025em b}\kern-.08em
    T\kern-.1667em\lower.7ex\hbox{E}\kern-.125emX}}

\DeclareMathOperator*{\argmax}{\arg\!\max}
\newcommand{\mathvc}[1]{\mbox{\boldmath$#1$}}
\newcommand{\smathvc}[1]{\mbox{\scriptsize \boldmath$#1$}}
\newcommand{\tmathvc}[1]{\mbox{\tiny \boldmath$#1$}}

\begin{document}

\title{ Semi-Supervised Few-Shot Classification \\ with Deep Invertible Hybrid Models }


\author{\IEEEauthorblockN{Yusuke Ohtsubo, Tetsu Matsukawa, Einoshin Suzuki  }
  \IEEEauthorblockA{\textit{Graduate School/Faculty of Information Science and Electrical Engineering,} \\
    \textit{Kyushu University} Fukuoka, Japan \\
fee3c.k@gmail.com,  \{matsukawa, suzuki\}@inf.kyushu-u.ac.jp}
}
\maketitle
\begin{abstract}
In this paper, we propose a deep invertible hybrid model which integrates discriminative and generative learning at a latent space level for semi-supervised few-shot classification.
Various tasks for classifying new species from image data can be modeled as a semi-supervised few-shot classification, which assumes a labeled and unlabeled training examples and a small support set of the target classes. Predicting target classes with a few support examples per class makes the learning task difficult for existing semi-supervised classification methods, including self-training, which iteratively estimates class labels of unlabeled training examples to learn a classifier for the training classes.
To exploit unlabeled training examples effectively, we adopt as the objective function the composite likelihood, which integrates discriminative and generative learning and suits better with deep neural networks than the parameter coupling prior, the other popular integrated learning approach.
In our proposed model, the discriminative and generative models are respectively Prototypical Networks, which have shown excellent performance in various kinds of few-shot learning, and Normalizing Flow a deep invertible model which returns the exact marginal likelihood unlike the other three major methods, i.e., VAE, GAN, and autoregressive model. 
Our main originality lies in our integration of these components at a latent space level, which is effective in preventing overfitting.
Experiments using mini-ImageNet and VGG-Face datasets show that our method outperforms self-training based Prototypical Networks.
\end{abstract}

\section{Introduction}\label{sec:intro}
Suppose you wish to classify images into five kinds of new species given five reference images for each new species and a large number of images of known species.  
Few-shot classification~\cite{feifei06, lake15, vinyals16, snell17} models this kind of task using a large training set which consists of labeled examples of known classes and a small support set which consists of labeled examples of the target classes.


\par

Due to the high cost of obtaining labeled examples, semi-supervised learning
exploits unlabeled examples in addition to labeled examples in the training set.
Semi-supervised few-shot classification \cite{ren18,li19}, which is an extension of the few-shot classification,
also takes unlabeled training examples.
Such unlabeled training examples are considered useful in extracting common features among different classes, e.g., unlabeled examples could contribute to discovering spines for better vertebrate classification.

Existing methods for semi-supervised few-shot classification \cite{ren18,li19} use such unlabeled examples in the framework of 
self-training \cite{rosenberg05}.
In self-training, a model is first learnt from labeled training examples and used to predict the class labels of the unlabeled training examples.
Then these unlabeled examples with their predicted class labels are added to the training set to learn the next model and this process is iterated.
If the class labels of the unlabeled training examples are correctly predicted, the next model is expected to be more accurate than the former one,
which has been proved by the experiments \cite{ren18}.
\par

However, the opposite case in which unlabeled examples decrease classification accuracy happens if the prediction is poor.
Another series of experiments \cite{ren18} show that such a decrease happens when the unlabeled examples belong to different classes from the labeled training classes.
This procedure stems from the fact that self-training methods are based on a discriminative model, which requires identical classes to labeled examples. Unfortunately, the assumption that the classes of the unlabeled examples are identical to those of the labeled examples does not always hold true.



To solve this problem, we introduce a generative model, which allows us to exploit unlabeled examples with no assumption of their classes, in addition to a discriminative model. 
We integrate these models at the latent space level in the framework of composite likelihood \cite{mccallum06, nalisnick19}. Concretely speaking, we use, as the discriminative model, Prototypical network \cite{snell17}, which can solve few-shot classification, and, as the generative model, Normalizing Flow \cite{dinh14,kingma18} a deep neural network capable of computing exact likelihood.
Another advantage is that our framework can be systematically described in the framework of the maximum likelihood principle, whereas the self-training~\cite{ren18} deviates from it.

Note that there is a conjecture that a hybrid model of discriminative and generative models exhibits better predictive performance when the labeled training examples are scarce \cite{ng02, liang08}. 
Nevertheless, this conjecture has not been confirmed on a practical problem such as few-shot classification. Recently, a hybrid model \cite{nalisnick19} based on Normalizing  Flow has been proposed. 
However, it uses a regression model as the discriminative model and is thus unable to cope with the few-shot classification problem directly. We expand this approach \cite{nalisnick19} to a semi-supervised few-shot classification model and confirm the advantage of the hybrid model on this problem.

Our contributions are summarized as follows: 
\begin{itemize}
  \item We construct semi-supervised few-shot classification model without assuming that unlabeled data share the same class with labeled data. 
  \item We expand a deep hybrid generative/discriminative model to a semi-supervised few-shot classification model.
  \item We experimentally show the intermediate model of the generative and discriminative models achieve higher predictive performance than each of the component models on the semi-supervised few-shot classification problem.
\end{itemize}\par

This paper is structured as follows. Section \ref{sec:relation} explains related works and background of this paper. 
Section \ref{sec:meth-tp} explains our target problem. 
Sections \ref{sec:meth} and \ref{sec:results} explain our proposed hybrid model and experiments, respectively. 
Section \ref{sec:discussion} concludes.

\section{Related Work}\label{sec:relation}

\subsection{Semi-supervised Few-shot Classification}
Fei-Fei et al.~\cite{feifei06} initiated few-shot learning, and after Lake et al.'s work~\cite{lake15}, which uses a hierarchical model, the researches on the task became much more active. 
Later, many methods based on neural networks are proposed. 
Such methods can be classified into initialization based methods~\cite{FinnAL17, Finn18} and distance metric learning based methods~\cite{vinyals16, snell17}.
The former methods obtain good initial model parameters using the training set so that they can effectively cope with a small test set~\cite{FinnAL17, Finn18}. 
The latter methods learn a distance metric between a pair of images and estimate the test class based on the distances to a small number of test examples, e.g., Matching Networks \cite{vinyals16}. 
Prototypical Networks \cite{snell17} belong to the latter kind of methods, as they learn a distance metric between a sample and a class prototype which is defined as the mean of examples belonging to the class.
Note that we can use another distance metric learning based method instead of the Prototypical Networks.

Semi-supervised methods can be classified into generative model based methods \cite{karaletsos16} and self-training based methods \cite{yarowsky95,rosenberg05, ren18, li19}.
OPBN \cite{karaletsos16} belongs to the former and can include unlabeled data in the likelihood function. 
It can handle only binary classification and maximizes the lower-bound likelihood because it cannot obtain the exact likelihood.
Even if its exact likelihood could be obtained, its extension to multiway classification
by using Prototypical Networks corresponds to a special case (pure generative model) of our proposed method.
Prototypical Networks with K-means (PNK) \cite{ren18} belongs to the self-training, which iteratively estimates the class labels of unlabeled examples to improve the test accuracy of the classifier.
In the self-training based methods \cite{ren18, li19}, unlabeled training examples which belong to different classes from the labeled training classes are considered to have no contribution in the improvement. 
In our experiments, we will compare the proposed method with PNK due to its simplicity for implementation.
\par


\subsection{Deep Generative Model}\label{sec:relation-gen}
A deep generative model obtains the marginal density with a neural network~\cite{salakhutdinov06}. It evolved into methods which efficiently estimate parameters with the backpropagation algorithm, e.g.,
VAE \cite{kingma14,rezende14}, GAN\cite{goodfellow14}, Autoregressive model \cite{graves13} and Flow \cite{dinh14}.
VAE \cite{kingma14,rezende14} maximizes the lower-bound of the log-likelihood, and it has computational challenges \cite{nalisnick19}.
Since GAN \cite{goodfellow14} minimizes the Jensen-Shannon divergence instead of maximizing the likelihood,
it is difficult to adapt it in the framework of composite likelihood maximization.
Autoregressive model~\cite{graves13}  computes likelihood by conditioning each dimension of the input space on other dimensions sequentially, and thus its computation cannot be parallelized. Consequently, it is time-inefficient in learning from high-dimensional data including images~\cite{kingma18}.
Normalizing Flow \cite{dinh14} can obtain exact log-likelihood and thus can be extended to a method which maximizes the composite likelihood.
Though we adopted Glow \cite{kingma18} as the architecture of Normalizing Flow, we can also adopt other kinds of architectures \cite{dinh14,dinh17} in our approach.
Refer to an excellent survey \cite{papamakarios19} of deep invertible models for details.

\par

\subsection{Generative and Discriminative Models }\label{sec:meth-abs}

\par


Let us denote \mathvc{x} an example and $y$ a label, $\mathcal{X}^{\rm L}$ labeled examples and $\mathcal{Y}$ its labels.  
Given a parametric model which is governed by model paramters \mathvc{\theta}, 
many methods in the maximum likelihood approach either assume a generative model
$p(y, \mathvc{x} |\mathvc{\theta})$, or a discriminative model $p(y| \mathvc{x}, \mathvc{\theta})$ as the likelihood function to estimate the maximum likelihood estimate $\hat{\mathvc{\theta}}_{\rm ML}$. 
Assuming a generative model allows us to include unlabeled data $\mathcal{X}^U$ in addition to the labeled data $\mathcal{X}^L, \mathcal{Y}$ in the likelihood function \cite{lassere06}, which results in
\begin{eqnarray}
\label{eq:meth-abs-gen}
\hat{\mathvc{\theta}}_{\rm ML}=\argmax_{\mathvc{\theta}} \; p(\mathcal{Y}, \mathcal{X}^{\rm L},\mathcal{X}^{\rm U}|{\mathvc{\theta}}).
\end{eqnarray}
Assuming a discriminative model results in $\hat{\mathvc{\theta}}_{\rm ML}=\argmax_{\smathvc{\theta}} p(\mathcal{Y}|\mathcal{X}^{\rm U}, \mathcal{X}^{\rm L}, \mathvc{\theta})$,
though in practice it is difficult to include unlabeled data $\mathcal{X}^{\rm U}$ in the likelihood function.
It is common to adopt the supervised learning setting and thus
\begin{eqnarray}
\label{eq:meth-abs-dis}
\hat{\mathvc{\theta}}_{\rm ML}=\argmax_{\mathvc{\theta}} \: p(\mathcal{Y}|\mathcal{X}^{\rm L}, \mathvc{\theta}).
\end{eqnarray}
\par

There are pros and cons for the discriminative model $p( y | \mathvc{x}, \mathvc{\theta})$ and the generative model $p( y, \mathvc{x} | \mathvc{\theta})$.
Vapnik \cite{vapnik98} stated ``one should solve the problem (such as modeling $p(y|\mathvc{x}, \mathvc{\theta})$) directly and never solve a more general problem (such as modeling $p(y,\mathvc{x})$) as an intermediate step".
On the other hand, Ng et al. \cite{ng02} proved the following theorem for linear models.
Though the optimal generalization loss, i.e., the generalization loss with infinite training data, is lower for discriminative models than generative models, the latter requires fewer data than the former to attain it.
A discriminative model yields a low generalization loss with abundant training data \cite{ng02}.
On the other hand, a generative model, which attains the optimal generalization loss with fewer data, happens to show a lower generalization loss than a discriminative model when the training data is scarce \cite{ng02}.
In other words, one should use a discriminative model when the training data are abundant but he/she might achieve a lower generalization loss with a generative model when the training data are scarce.
In addition, a generative model can exploit unlabeled data in the likelihood function~\cite{lassere06} as shown in Eq.(\ref{eq:meth-abs-gen}).

\subsection{Hybrid Models}
To integrate generative and discriminative models, parameter coupling prior \cite{lassere06} and composite likelihood \cite{mccallum06} are proposed.
The parameter coupling prior uses separate model parameters for generative and discriminate models and considers their coupled prior distribution. 
The coupled prior distribution regularizes training such that parameters of generative and discriminative models become similar by means of Euclidean distance. 
We will use the neural networks for the modeling of the likelihood. 
In this case, the Euclidean distance between the parameters does not measure valuable differences between the functions realized by the neural networks \cite{tulyakov17}.

The composite likelihood, which is also known as pseudo-likelihood \cite{liang08} or multi-conditional learning \cite{mccallum06}, represents a weighted product of component likelihood \cite{liang08} and is better suited for integrating the likelihoods modeled by neural networks.
Liang et al conducted a theoretical analysis of the behavior of the composite model. 
Hybrid models are extended to neural network based methods \cite{shu17,kuleshov17}. 
Several works maximize lower-bounds of the log-likelihood of the model using stochastic gradient variational inference \cite{kingma14,rezende14}. 
Recently, a more direct method \cite{nalisnick19}, which maximizes the composite likelihood by using Normalizing Flow \cite{dinh14}, a neural network capable of conducting an inverse transformation from the latent space to the input space, has been proposed. 
However, it uses a regression model as the discriminative model and is thus unable to cope with new test classes in the few-shot classification problem. 
We expand this approach \cite{nalisnick19} to a distance metric-based semi-supervised few-shot classification model.


\section{Target Problem}\label{sec:meth-tp}
\begin{figure}[tbp]
 \centering
 \includegraphics[keepaspectratio, width=\linewidth]
      {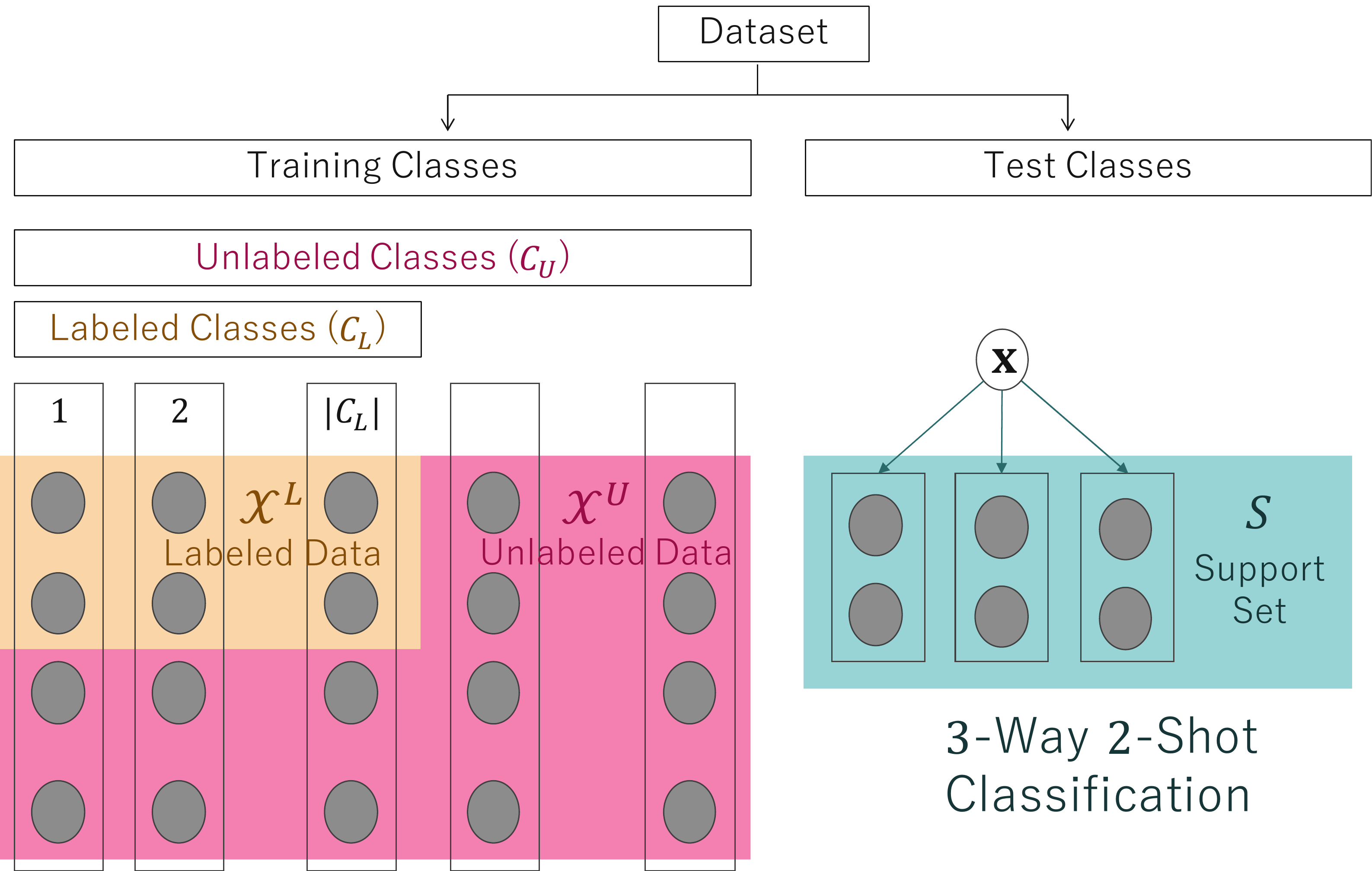}
 \caption{Semi-supervised few-shot classification problem}
 \label{fig:meth-ssfs}
\end{figure}

Figure \ref{fig:meth-ssfs} shows the semi-supervised few-shot classification problem.
The task is to classify a test example \mathvc{x} into one of the test classes, given a large training set $ \mathcal{D}^{\rm train} = \{\mathcal{X}^{\rm L}, \mathcal{Y}, \mathcal{X}^{\rm U}\}$ and a support set $\mathcal{S}$. 
We refer to the classes included in the training and support sets as training and test classes, respectively.

The training set consists of a set of $L$ labeled examples, $\{\mathcal{X}^{\rm L}, \mathcal{Y} \} =\{(\mathvc{x}_{l}, y_l)\}_{l=1}^{L}$ and a set of $U$ unlabeled examples $\mathcal{X}^{\rm U}=\{\mathvc{x}_{u}\}_{u=1}^{U} $ where two kinds of training examples $\mathvc{x}_l, \mathvc{x}_u$ are in an identical input space\footnotemark~ and $y_l$ is a class label in labeled training classes $\mathcal{C}_{\rm L}$. 
\footnotetext{ In this paper, the input space is image space, and we model the mapping to a latent space by a deep neural network. }
It is common to assume that each unlabeled example $\mathvc{x}_u$ belongs to one of (unknown) unlabeled classes $\mathcal{C}_{\rm U}$.
Usually, the unlabeled classes $\mathcal{C}_{\rm U}$ include several classes of $\mathcal{C}_{\rm L}$.

The support set $\mathcal{S}$ consists of labeled examples which are described in the same input space and belonging to test class labels. The test classes are different from both the training classes $\mathcal{C}_{\rm L}$ and $\mathcal{C}_{\rm U}$.
We tackle $C$-way $K$-shot classification, in which the number of the test classes is $C$, and the support set contains $K$ examples for each of the test classes.

\section{Proposed Model}\label{sec:meth}





\subsection{ Our Model }\label{sec:meth-abs}

From a probabilistic perspective, we model the few-shot classification and estimate a predictive distribution of label $y$ over the $C$ test classes for each test example \mathvc{x} given the training and support sets. 
In case of the Bayesian inference, a predictive distribution of $y$ is given by $p(y|{\mathvc{x}}, \mathcal{S}, \mathcal{D}^{\rm train})$. 
We introduce a parametric model which is governed by model parameters \mathvc{\theta}. 
In this paper, we conduct maximum likelihood estimation on the training set and thus the predictive distribution becomes $p(y|{\mathvc{x}}, \mathcal{S}, \hat{\mathvc{\theta}}_{\rm ML})$  in inference time. 

In the training, we make use of the composite likelihood model~\cite{liang08, mccallum06} which is a weighted product of a discriminative model and a generative model.
The composite likelihood is given as:
\begin{eqnarray}
\label{eq:meth-comp}
\mathcal{L}(\alpha, \beta) = p(\mathcal{Y}|\mathcal{X}^{\rm L}, \mathvc{\theta})^\alpha \; p(\mathcal{X}^{\rm L},\mathcal{X}^{\rm U}| \mathvc{\theta})^\beta,
\end{eqnarray}
where $\alpha\geq 0$ and $\beta\geq 0$ are the weights of the discriminative model and the weight of the marginal likelihood of 
\mathvc{x}, respectively.
Note that $\mathcal{L}(1, 0)$ and $\mathcal{L}(1, 1)$ represent the likelihood functions of
a pure discriminative and generative models, respectively. Since the influences of $\alpha$ and $\beta$ are relative in a maximum likelihood estimation, for simplicity we set $\alpha=1$ and take $\beta$ as a variable.
Thus we hereafter denote $\mathcal{L}(1,\beta)$ by $\mathcal{L}$.\par


We introduce a latent space realized with an invertible neural network \cite{dinh14} for modeling the marginal likelihood and the discriminative model. 
We assume a latent space which is shared by the labeled and unlabeled training examples. The shared latent space allows us to exploit the unlabeled training examples regardless of their classes.

In the following, Section \ref{sec:meth-gen} explains the marginal likelihood $p(\mathcal{X}^{\rm L}, \mathcal{X}^{\rm U}| \mathvc{\theta})$ which is a component of the 
composite likelihood.
Section \ref{sec:meth-dis} explains the discriminative model $p(\mathcal{Y}|\mathcal{X}^{\rm L}, \mathvc{\theta})$. 
Section \ref{sec:meth-hybrid} explains the resulting hybrid model.
Section \ref{sec:meth-opt} explains the details of the paramter estimation. 


\subsection{Marginal Density Model: Normalizing Flow}\label{sec:meth-gen}

For our marginal density model, we use Normalizing Flow \cite{dinh14} a deep generative model which can efficiently obtain marginal density\footnote{We described the reason other deep generative models are inadequate as a component of the composite likelihood in Section \ref{sec:relation-gen}.}
$p(\mathcal{X}^{\rm L},\mathcal{X}^{\rm U}|\mathvc{\theta})$.
We assume that the examples are mutually independent, i.e., 
$p(\mathcal{X}^{\rm L},\mathcal{X}^{\rm U}| \mathvc{\theta}) = \prod_{l=1}^L p( \mathvc{x}_l | \mathvc{\theta}) \prod_{u=1}^U p( \mathvc{x}_u | \mathvc{\theta}) = \prod_{n=1}^{L+U}p( \mathvc{x}_n | \mathvc{\theta})$ where $\{\mathvc{x}_n\}_{n=1}^L = \{\mathvc{x}_l\}_{l=1}^{L}$ and  $\{\mathvc{x}_n\}_{n=L+1}^{L+U} = \{\mathvc{x}_u\}_{u=1}^{U}$.
Normalizing Flow introduces latent variables $\mathvc{z}$ and assumes a simple distribution such as normal distribution as a prior distribution of the latent variable $p_z(\mathvc{z})$. 
It models complex\footnotemark~marginal density $p(\mathvc{x}|\mathvc{\theta})$ with a transformation using a neural network $\mathvc{x} = f_{\smathvc{\theta}}(\mathvc{z})$. 
Here, the network is composed of an architecture which allows the inverse transformation $\mathvc{z} = f^{-1}_{\smathvc{\theta}}(\mathvc{x})$.
\footnotetext{For instance, it can represent complex natural images.}
Under the transformation of variables by $\mathvc{x} = f_{\smathvc{\theta}}(\mathvc{z})$, marginal density $p(\mathvc{x}|\mathvc{\theta})$ is given by the following equation:
\begin{eqnarray}
\label{eq:meth-Flow}
p(\mathvc{x}_n|\mathvc{\theta})=p_z(f_{\smathvc{\theta}}^{-1}(\mathvc{x}_n)) \left| {\rm det} \left( \frac{\partial f_{\smathvc{\theta}}^{-1}({\mathvc{x}_n})}{\partial \mathvc{x}_n} \right) \right|,
\end{eqnarray}
where $\left|{\rm det} \left(\frac{\partial f_{\tmathvc{\theta}}^{-1}(\mathvc{x})}{\partial \mathvc{x}} \right) \right|$
represents the absolute value of the Jacobian determinant of the transformation $f^{-1}_{\smathvc{\theta}}(\cdot)$ at \mathvc{x}. 

Real-valued Non-Volume Preserving (RNVP) transformation \cite{dinh17} and Glow transformation \cite{kingma18} are popular architectures
of neural networks $f_{\smathvc{\theta}}(\cdot)$ which allow the inverse transformation. We used the Glow architecture for our experiments. For the details of its architecture, see \cite{kingma18}.
The crux here is that by the introduction of a neural network $\mathvc{x}=f_{\smathvc{\theta}}(\mathvc{z})$ which allows inverse transformation, 
we can obtain the gradient of the exact marginal likelihood $\frac{\partial p(\mathvc{x}|\smathvc{\theta})}{\partial \smathvc{\theta}}$,
which is necessary to optimize the model parameters in the composite likelihood of Eq.~(\ref{eq:meth-comp}).

\subsection{Discriminative Model: Prototypical Networks}\label{sec:meth-dis}
Deep Invertible Generalized Linear Model (DIGLM) \cite{nalisnick19} also uses Normalizing Flow in the framework of the composite likelihood model. 
However, it uses a regression model as the discriminative model
$p(\mathcal{Y}|\mathcal{X}^L, \mathvc{\theta})$ and is thus unable to cope with
new test classes in the few-shot classification problem.
As we explained in Section \ref{sec:relation},
there are several works which use neural networks to tackle few-shot classification.
Prototypical Networks~\cite{snell17} measure the distance between an example $\mathvc{x}$ and the prototype of each class and assign a higher probability $p(y | \mathvc{x}, \mathvc{\theta})$ to a closer class to $\mathvc{x}$.
We adopt this model, which despite its simplicity, shows high accuracy in experiments.
They also enable easy comparison with a self-training based method\cite{ren18}.
\par

For optimizing model parameters \mathvc{\theta} from $\mathcal{X}^{\rm L}$, we employ Stochastic Gradient Descend (SGD) for efficiency \cite{snell17}. 
To form the prototype of each class, a sample set $\mathcal{X}_l^\mathcal{S}$ is randomly chosen from the labeled training set for each example $\mathvc{x}_l$.
It is common to form the sample set by mimicking the support set, i.e.,
the sample set consists of $CK$ examples belonging to $C$ classes,
in which the number of examples for each class is $K$.
The pair of the test example
$\mathvc{x}_l$ and the corresponding sample set $\mathcal{X}^\mathcal{S}_l$, which corresponds to a minibatch in SGD, is
called an episode \cite{snell17}.
\par
We define a prototype in the latent space obtained by the inverse transformation $\mathvc{z}_l=f^{-1}_{\smathvc{\theta}}(\mathvc{x}_l)$ of Normalizing Flow.
For each example $\mathvc{x}_l$, the prototype $\mathvc{t}^c_l$ of class $c$
is defined as the mean of the examples in class $c$ in the latent variable set $\mathcal{Z}_l^\mathcal{S}$
of the sample set as follows.
\begin{eqnarray}
\label{eq:meth-dis-proto}
\mathvc{t}_l^c = \frac{1}{|\mathcal{Z}_l^{\mathcal{S}_c}|}\sum_{\mathvc{z}_i\in \mathcal{Z}_l^{\mathcal{S}_c}} \mathvc{z}_i. 
\end{eqnarray}
The probability that example $\mathvc{x}_l$ belongs to class $c$ is defined with a softmax function
which takes the distance to each class prototype as the input.
\begin{eqnarray}
\label{eq:meth-dis-pred}
p(y_l=c| \mathvc{z}_l, \mathcal{Z}_l^{\mathcal{S}})=\frac{\exp(-\|\mathvc{z}_l-\mathvc{t}_l^c\|^2_2)}{\sum_{c'}\exp(-\| \mathvc{z}_l - \mathvc{t}_l^{c'}\|^2_2)}.
\end{eqnarray}

\begin{figure}[t]
        \centering
         \includegraphics[width=0.8 \linewidth]{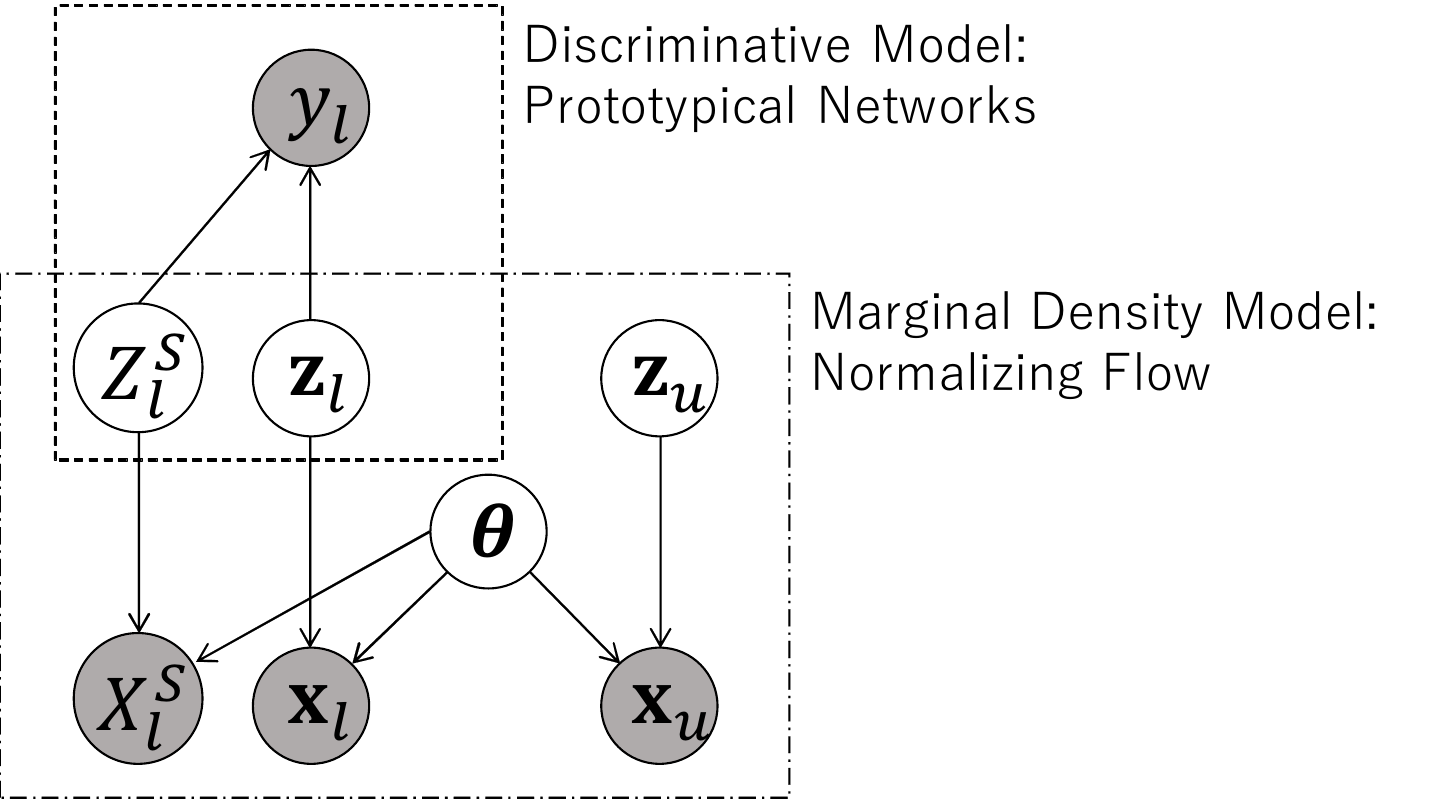}
\caption{Graphical model of the proposed method. 
The lower dashed rectangle is the marginal likelihood $p(\mathcal{X}^{\rm L}, \mathcal{X}^{\rm U}| \mathvc{\theta})$ explained in Section \ref{sec:meth-gen}.
The upper dashed rectangle is the discriminative model $p(\mathcal{Y}|\mathcal{X}^{\rm L}, \mathvc{\theta})$ explained in Section \ref{sec:meth-dis}.
The labeled example $\mathvc{x}_l$ and its label $y_l$, the sample set $\mathcal{X}_l^\mathcal{S}$, and the unlabeled example $\mathvc{x}_u$ are half-toned as observed variables.
The latent variable $\mathvc{z}$ takes the roles of both image generation and label estimation. Since parameters $\mathvc{\theta}$ are shared between the labeled and unlabeled data, the unlabeled data contribute to the estimation of the latent variables for the labeled data.}
 \label{fig:prop-graphical}
\end{figure}

\subsection{Hybrid Model}\label{sec:meth-hybrid}

By introducing the marginal density and discriminative models and taking logarithm to Eq.~(\ref{eq:meth-comp}), we obtain the following log composite likelihood function that our hybrid model maximizes:
\begin{eqnarray}
{\rm log} \mathcal{L}
=\sum_{l=1}^{L}\log p(y_l| \mathvc{z}_l, \mathcal{Z}_l^{\mathcal{S}})+\beta\sum_{n=1}^{L+U} \log p(\mathvc{x}_n|\mathvc{\theta}). \label{eq:meth-hybrid-low}
\end{eqnarray}
The first term represents the log-likehoood of the discriminative model based on Eq.~(\ref{eq:meth-dis-pred}), while the second term represents that of the marginal density based on Eq.~(\ref{eq:meth-Flow}).

Figure \ref{fig:prop-graphical} shows our graphical model, which integrates graphical models of Normalizing Flow~\cite{kingma18} and Prototypical Networks~\cite{snell17} for our hybrid model. 
First, note that labeled example $\mathvc{x}_l$ is generated by latent variables $\mathvc{z}_l$, which is realized by the Normalizing Flow neural network with parameters $\mathvc{\theta}$.
Unlabeled example $\mathvc{x}_u$ is generated similarly.
From these facts we can expect that the shared parameters $\mathvc{\theta}$
are learnt such that they express the essential characteristics of $\mathvc{x}_u$
and concurrently $\mathvc{\theta}$ are used to estimate $\mathvc{z}_l$ from labeled data
$\mathvc{x}_l$.
Our advantage over Ren et al.'s self-training method \cite{ren18} lies in the fact that the above process is conducted with no assumption on the classes of the unlabeled
training examples.

Second, note that the process of estimating class $y_l$ of $\mathvc{z}_l$
from the relationship between latent variables $\mathvc{z}_l$ and set $\mathcal{Z}_l^S$ of the latent variables in the
support set.
Together with the fact that labeled example $\mathvc{x}_l$ is generated from 
$\mathvc{z}_l$, we can expect that 
$\mathvc{z}_l$ takes the trade-off of the discriminative characteristics for estimating $y_l$ and the
generative characteristics for generating $\mathvc{x}_l$ \cite{ng02}\footnote{The generative learning is called informative learning in \cite{rubinstein97}.}
and $\beta$ balances the two kinds of characteristics.
Note that Eq.~(\ref{eq:meth-hybrid-low}) shows that our method degenerates to Prototypical Networks \cite{snell17} of Eq.~(\ref{eq:meth-dis-pred})
when $\beta=0$. When $\beta=1$, it degenerates to a pure generative model (or the joint model).
\par

\subsection{Optimization}\label{sec:meth-opt}

Since the optimization of $\beta$ should be based on the minimization principle of the generalization loss, we could use WAIC \cite{watanabe10}, which approximates the generalization loss, or a minimization by the leave-one-out cross validation.
However, WAIC requires sampling from the posterior distribution of the parameters,
which is unknown for Normalizing Flow. 
Leave-one-out cross validation is notoriously slow when the training data is large.
Thus in the experiments we optimize $\beta$ based on the generalization loss on
a validation set, which is kept apart from the training and test sets.

Parameters $\mathvc{\theta}$ are updated based on the backpropagation algorithm.
For a faster learning with less memory consumption, we update the parameters with SGD, in which the log-likelihood function in Eq.~(\ref{eq:meth-hybrid-low}) is iteratively maximized in terms of the minibatch.
For the term of the discriminative model, we form the sample set, i.e., the labeled data in the mini-batch, by $CK$ examples from the labeled training data as explained in Section \ref{sec:meth-dis}. We firstly sample $C$ classes from the training classes and then selected $K$ examples randomly for each class. For the marginal density term, we sample unlabeled training data randomly, which reflects that their class labels are unknown. The ratio between the labeled and unlabeled examples in a minibatch is identical to that in the whole training data so that the objective function on the minibatch well approximates that on the entire training data.
\par


\section{Experiments}\label{sec:results}
In this Section, we evaluate our hybrid model on two datasets, i.e., mini-ImageNet \cite{vinyals16} and VGG-Face \cite{parkhi15}.
We first explain the details of our implementation in Section \ref{sec:exp-imp} and then 
the datasets and the experimental settings in Section \ref{sec:exp-data}. 
To validate our modeling, which has no assumption on the classes of the unlabeled examples, we evaluate their predictive distributions by changing the number of the unlabeled training examples. Section \ref{sec:exp-adduc1} evaluates them when none of the unlabeled training examples belong to the labeled classes.  
Section \ref{sec:exp-adduc2} evaluates them when some of the unlabeled training examples belong to labeled classes. 
Moreover, we measure the influence of weight $\beta$ for the generative model in Section \ref{sec:exp-beta}.

\subsection{Implementation Details}\label{sec:exp-imp}

\noindent {\bf Network architecture.} 
We adopted a Glow \cite{kingma18} architecture, a kind of a Normalizing Flow architecture, with $4$-blocks of $4$-affine coupling layers. 
We resized each input image using Lanczos function\footnotemark~and normalized each RGB value so that their means and their variances become $0$ and $1$, respectively.
\footnotetext{The size of resized image is $80*80*3$ and $64*64*3$ pixels, respectively, for the mini-ImageNet and the VGG-Face datasets.}
We used the normalized image as the input of our Glow architecture.
We introduced convolutional layers because their effectiveness in the Glow architecture on image data has been confirmed by experiments~\cite{kingma18}.
We had to use a shallower structure than the original implementation ($3$-blocks of  $32$-affine coupling layers) \cite{kingma18} due to our memory constraint. 
Adopting a public implementation~\cite{nalisnick19b}, we constructed an affine coupling layer by a $3$-layer Highway network with $216$ hidden units.
\par

\noindent {\bf Baseline method.} 
We adopted a self-training based method Prototypical Networks with soft K-means (PNK) \cite{ren18} due to its simplicity.
PNK adopts two strategies in disregarding unlabeled training examples which are
estimated not to belong to the training classes.
One strategy is to classify them into a distractor cluster and the other one is to
decrease their  influence by masking. Since masking exhibits better performance
in most of the experiments by Ren et al.~\cite{ren18}, we adopt it in our baseline method.
We modified the network structure of PNK to the Glow architecture with $4$-blocks of $4$-affine coupling layers, which our method adopts. 
Though our baseline method is not identical to the corresponding one in \cite{ren18},
this modification allows us to evaluate the differences of the two objective functions.
Note that we adopted only PNK with masking as the baseline method due to the high cost of re-implementing other methods with Glow.
\par

\noindent {\bf Parameter setting.}
We used ADAM\cite{kingma15} with the the learning rate $5*10^{-4}$, and hyperparramers $\beta_1=0.9$, and $\beta_2=0.999$.
We set the weight decay to $5*10^{-5}$.
These four hyper parameters were fixed before the execution and were not tuned with the validation set.



We conducted a grid search on the validation set to adopt the pair of the weight $\beta \in \{10^{-4}, 10^{-5}, ..., 10^{-9}\}$ and the number of epochs $\in \{ 1,2, ...,80 \}$ that achieves the highest evaluation measures.
We conduct the parameter search for each of the different settings of the training samples.
Similarly, we determined the number of epochs from the same range for the baseline method.

\subsection{Datasets and Settings}\label{sec:exp-data}
\noindent {\bf mini-ImageNet}~\cite{vinyals16}. The dataset is an edited version of ILSVRC-12 dataset~\cite{ILSVRC15}. 
It consists of $60,000$ images belonging to $100$ classes, each of which has $600$ images. We followed the protocol in \cite{ren18} and assigned $64$ training classes, $16$ validation classes, and $20$ test classes.
Among the 64 training classes, $50$ classes were selected as labeled classes $\mathcal{C}_{\rm L}$ and $14$ classes were hidden to generate the unlabeled training examples.
For the labeled training examples $\mathcal{X}^{\rm L}$, we selected $40\%$ of examples, i.e., $240$ examples, per each labeled class.


\noindent {\bf VGG-Face}~\cite{parkhi15}. 
The dataset consists of facial images of 2,622 persons, each of whom corresponds to a class. 
Among the persons, 2,353 persons contain no less than 100 images, and we used them in the experiments. We randomly selected 100 images for each person.
The $2,353$ classes were divided into $1800$ training classes, $200$ validation classes, and $353$ test classes.
In the training classes, $300$ classes were selected as labeled classes $\mathcal{C}_{L}$ and $1500$ classes were hidden to generate the unlabeled training examples. 
As the labeled training examples $\mathcal{X}^{\rm L}$, we selected $10\%$ of examples, i.e., $10$ examples, per each labeled class.

\par
Note that we change the setup of unlabeled examples $\mathcal{X}^{\rm U}$ and classes $\mathcal{C}_{\rm U}$ in the following experiments in Sections \ref{sec:exp-adduc1}, \ref{sec:exp-adduc2}, and \ref{sec:exp-beta}, while the labeled training examples $\mathcal{X}^{\rm L}$ are common for all experiments.


\noindent {\bf Evaluation.} 
To form the support set in the test set in $C$-way $K$-shot classification,
for each test example \mathvc{x}, we randomly selected $C-1$ classes from the test classes,
and then randomly selected  $K$ examples for each class\footnote{In the process we did not include the test example \mathvc{x}.}.
We kept the condition $C=5$ and $K=5$ throughout the experiments. 

We adopted as the evaluation measure the test accuracy and the negative cross entropy (the higher, the better) between the estimated class distribution and the true class labels~\footnote{The negative cross-entropy is also measured on the test examples. }.
Note that the accuracy, despite its clear meaning, is only based on the mode of the estimated class distribution.
In contrast, the negative cross entropy directly measures the estimated class distribution.
We believe that the latter measure is also valuable because it is more informative and thus gives a basis for further detailed analyses.

\subsection{Case When the Unlabeled and Labeled Training Classes are Disjoint}\label{sec:exp-adduc1}
First, we evaluate our method when the labeled and unlabeled classes are disjoint from each other, i.e., $\mathcal{C}_{\rm U}\cap \mathcal{C}_{\rm L} = \emptyset$. 
In this case, the assumption on the unlabeled examples of the baseline method~\cite{ren18} is not satisfied. 
We validate our hypothesis that our method improves predictive distributions even in such a case.

We construct unlabeled training examples $\mathcal{X}^{\rm U}$ from the training examples which do not belong to labeled classes. As the unlabeled classes $\mathcal{C}_{\rm U}$, we used $\{0, 2, 6, 10, 14\}$ classes, each of which having $600$ examples for the mini-ImageNet dataset. 
For the VGG-Face dataset, we used $\{0, 300, 900, 1500\}$ classes, each of which having $10$ examples. 


\def\subfigcapskip{-1.5pt}  
\begin{figure*}[tbp]
     \begin{subfigure}[{\small mini-ImageNet (left: accuracy, right: negative cross entropy)}]
                {\includegraphics[width=0.245\linewidth]{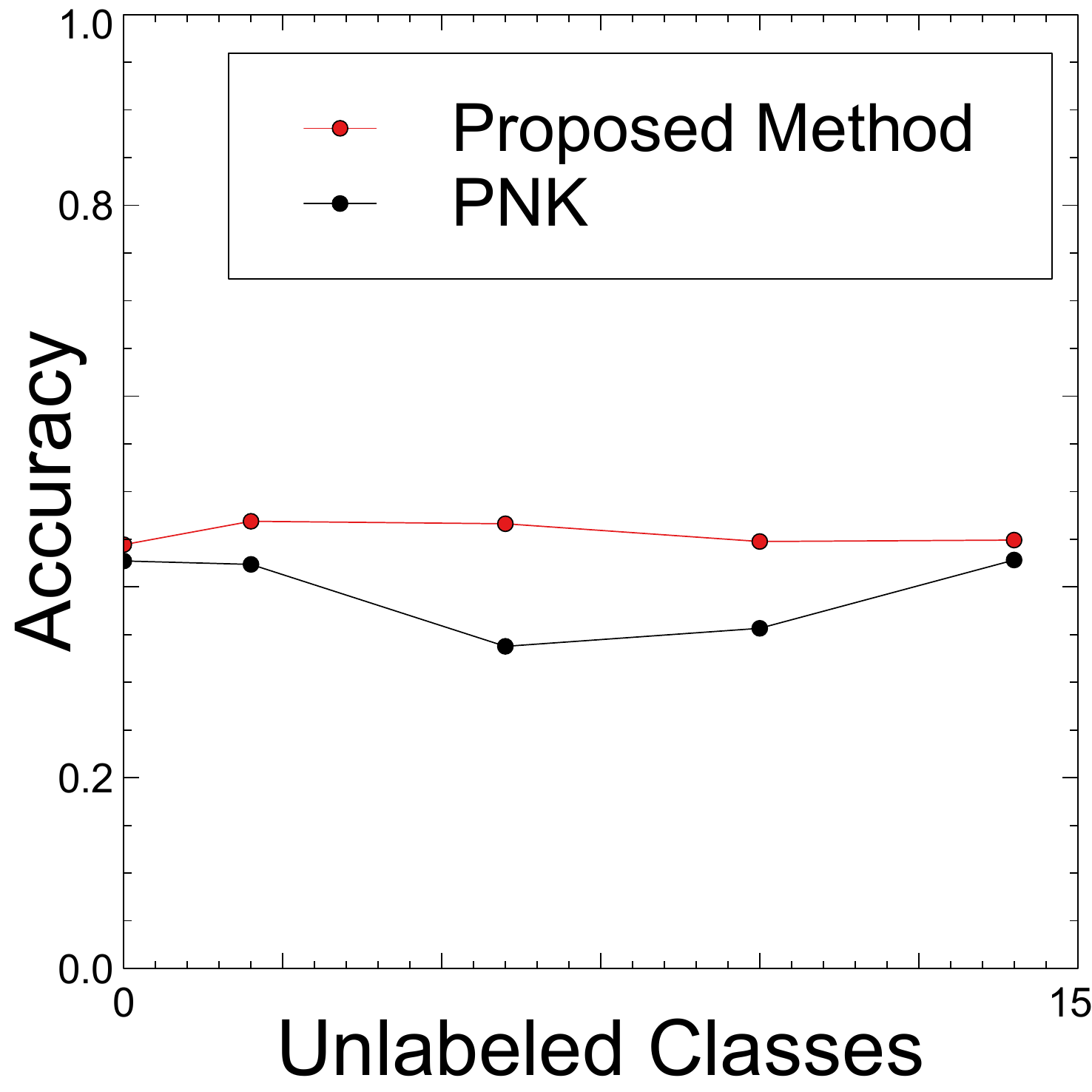}
                \includegraphics[width=0.245\linewidth]{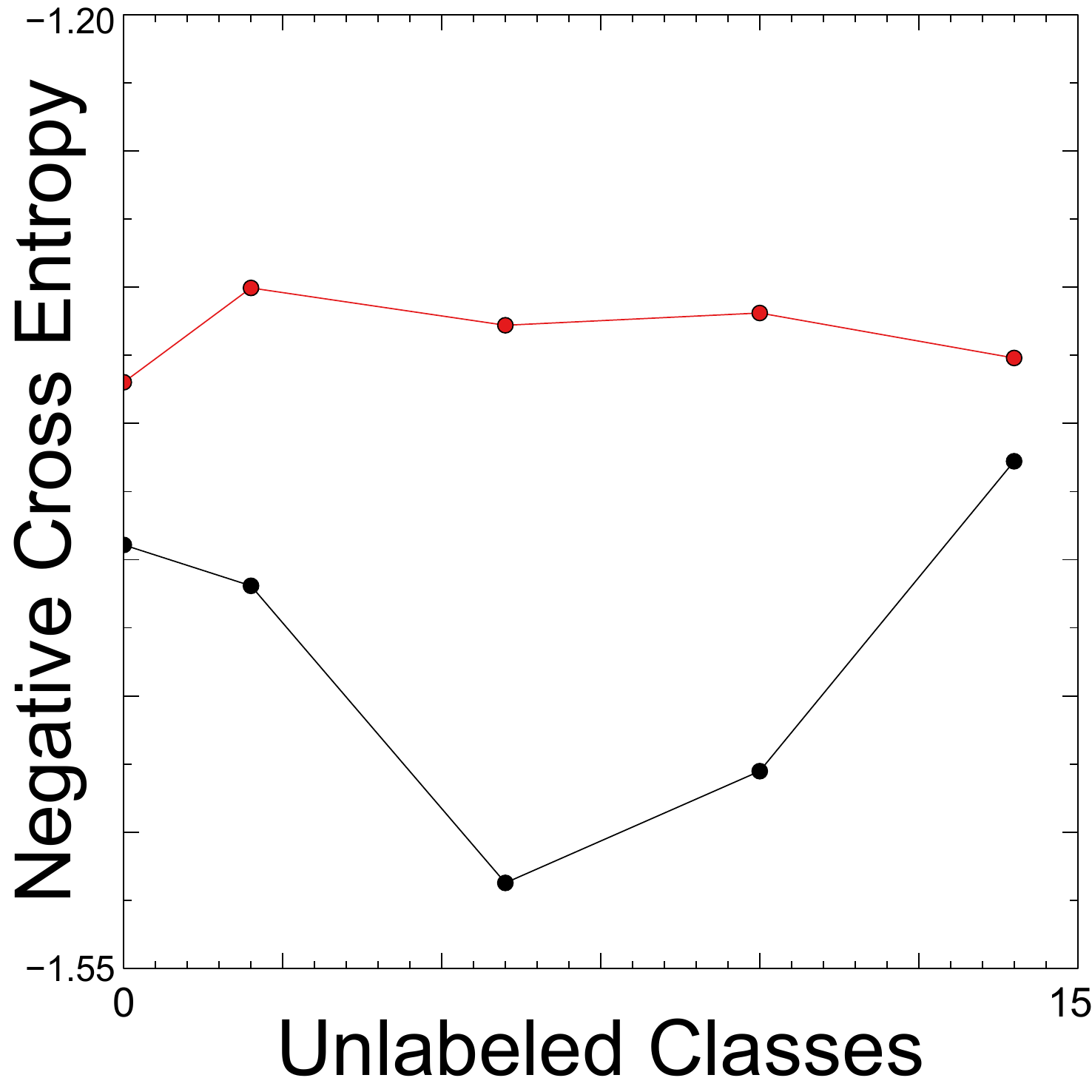}
                }
        \end{subfigure}%
        \begin{subfigure}[{\small VGG-Face (left: accuracy, right: negative cross entropy)}]
                {\includegraphics[width=0.245\linewidth]{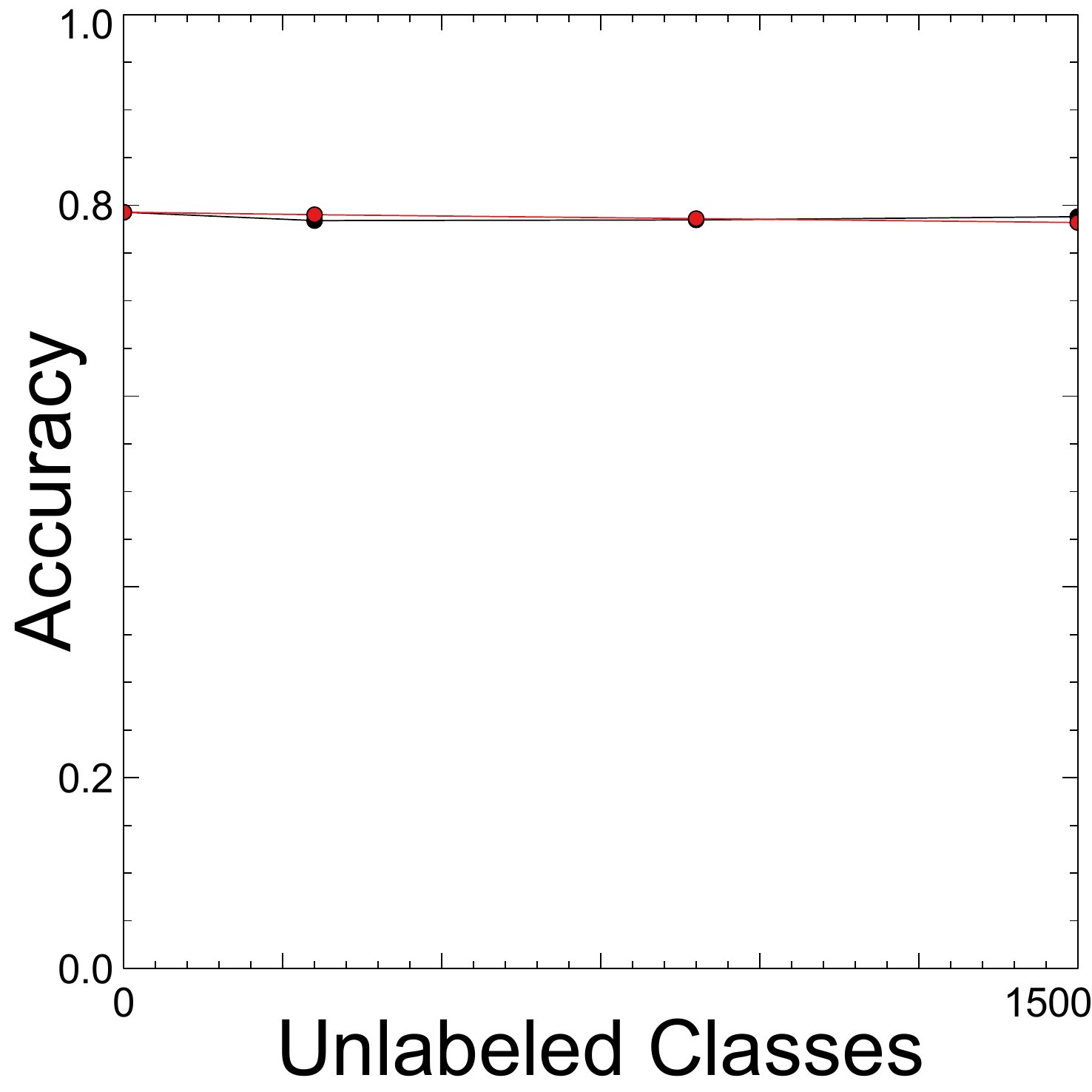}
                \includegraphics[width=0.245\linewidth]{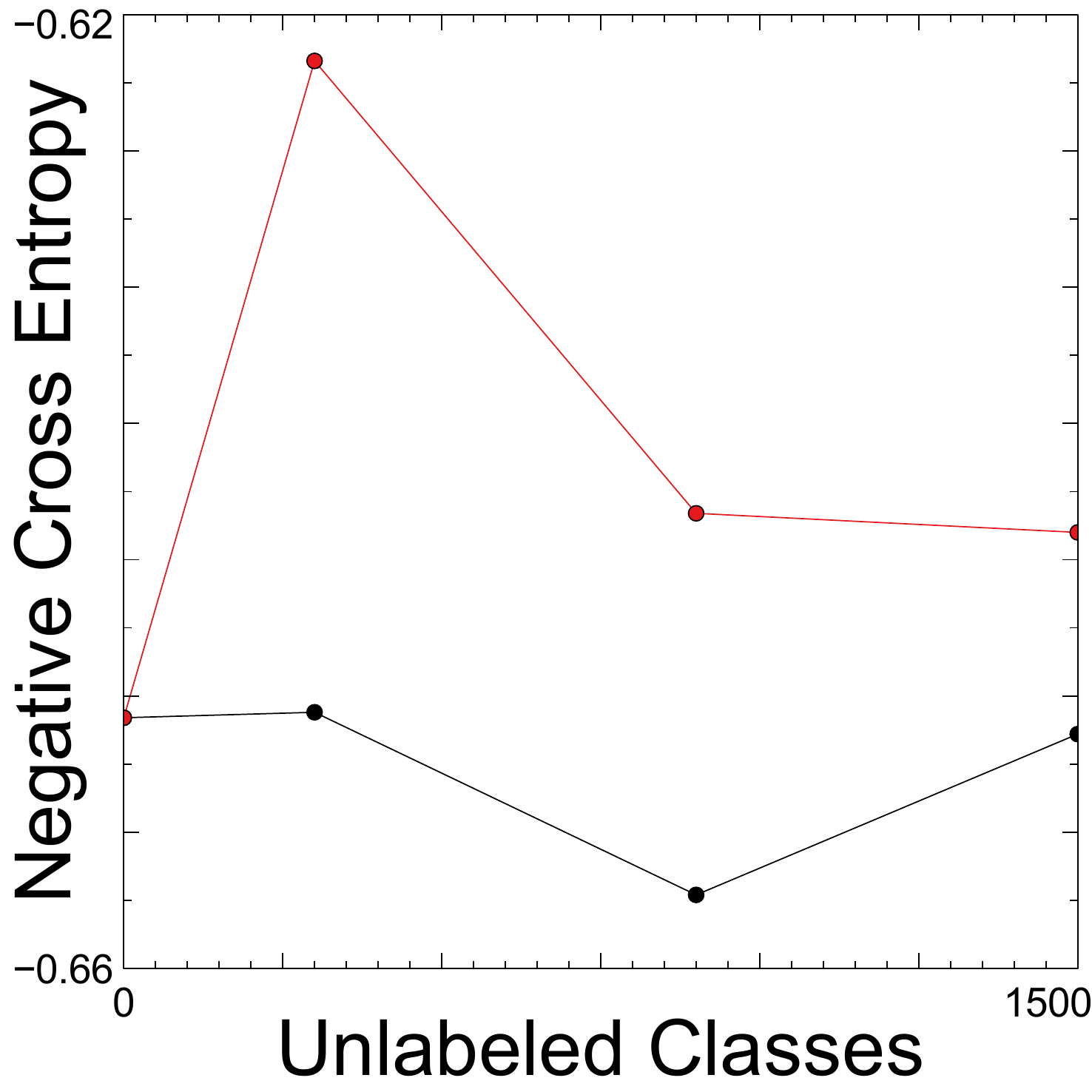}
                }
        \end{subfigure}%

       \vspace{-2mm}
        \caption{Predictive performance with increasing unlabeled classes when $\mathcal{C}_{\rm U}\cap \mathcal{C}_{\rm L} = \emptyset$ 
}\label{fig:exp-us0} 
\end{figure*}

\par
\subsubsection{Results on mini-ImageNet}
Figure \ref{fig:exp-us0} (a) shows the results on the mini-ImageNet dataset. In the Figure, the red and black curves represent our proposed method and the baseline method \cite{ren18}, respectively. The x-axis of each plot represents the number $|C_{U}|$ of the unlabeled classes, and the left endpoint represents the case with no unlabeled training example, i.e., supervised learning.
We see that the accuracies of the two methods are almost the same at the left endpoint. With an increasing number of classes of the unlabeled training examples, the accuracy and the negative cross entropy of the baseline method tend to decrease. In contrast, those of our method are stable or slightly increase.
These results support our hypothesis that the performance of our method improves or is at least stable by increasing number of the unlabeled training examples even when their underlying classes are different from the labeled training classes. The reason would be that our proposed method models the latent variable of the unlabeled training examples with no assumption to their classes (Fig. \ref{fig:prop-graphical}).
\par
\subsubsection{Results on VGG-Face}
Figure \ref{fig:exp-us0} (b) shows the results on the VGG-Face dataset. 
We see that the accuracies are almost the same for different numbers of the unlabeled training classes for both methods.
The reason would lie in the fact that the training and the test classes are highly similar, and thus the required number of the unlabeled examples is smaller than in the mini-ImageNet dataset.
From the left endpoint of Fig. \ref{fig:exp-us0} (b-right),  we see that the negative cross entropies of the two methods are the same when there is no unlabeled training example.
These results are not surprising because $\beta$ was set to 0 by the parameter setting on the validation set in our method, and as a result, both methods became pure Prototypical Networks.
Nevertheless, with the increasing number of classes of the unlabeled training data, the negative cross entropy of the baseline method deteriorates, whereas that of our method improves. These results again support our hypothesis as in the mini-ImageNet dataset.


\subsection{Case When the Unlabeled and Labeled Training Classes Overlap}\label{sec:exp-adduc2}
\noindent

The experimental condition in Section \ref{sec:exp-adduc1} are in some sense unrealistic because the classes of the unlabeled and labelled training examples are disjoint. 
As a more realistic setting, we evaluate our method in the case when the classes of the unlabeled and labeled training examples overlap, i.e., $\mathcal{C}_{U}\cap \mathcal{C}_{L} \neq \emptyset$.
We verify our hypothesis that the performance of our method improves or is at least stable in such a case.

For the mini-ImageNet dataset, as the unlabeled training set $\mathcal{X}^{\rm U}$, we first used the remaining $60\%$ of examples, i.e., 360 examples, from each of the $50$ labeled classes, i.e., $\mathcal{C}_{\rm U}=\mathcal{C}_{\rm L}$.
Then we added $\{0, 2, 6, 10, 14\}$ classes which are not included in $\mathcal{C}_{\rm L}$. We used $600$ examples per added class.   
Similarly, for the VGG-Face dataset, as unlabeled training examples $\mathcal{X}^{\rm U}$, we first used the remaining $20\%$ of the examples, i.e., 20 examples, from each of the $300$ labeled classes.
Then we added $\{0, 300, 900, 1500\}$ classes which are not included in the labeled classes. We used $10$ examples per added class. 

\begin{figure*}[tbp]
        \begin{subfigure}[{\small mini-ImageNet (left: accuracy, right: negative cross entropy)}]
                {\includegraphics[width=0.245\linewidth]{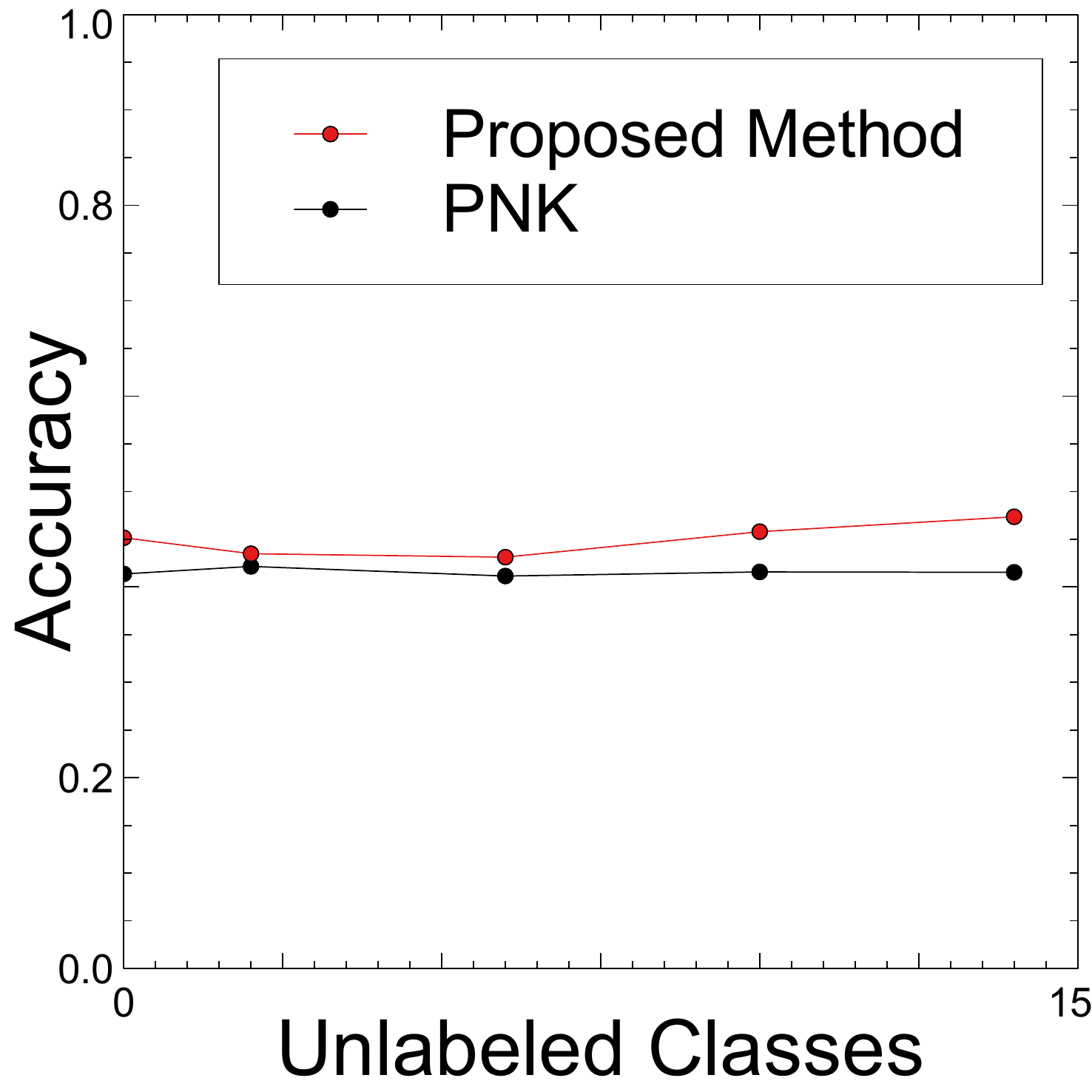}
                \includegraphics[width=0.245\linewidth]{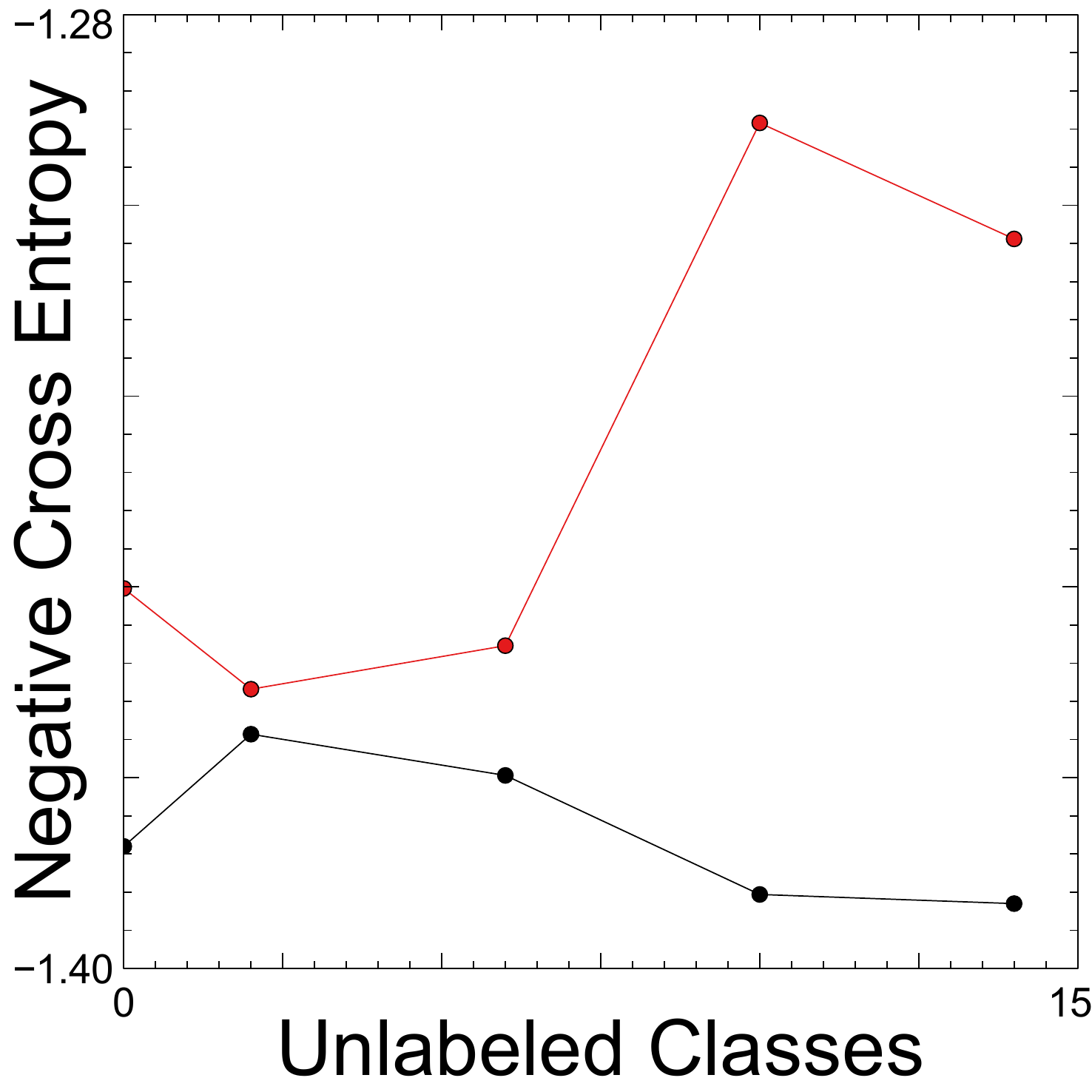}
                }
        \end{subfigure}%
        \begin{subfigure}[{\small VGG-Face (left: accuracy, right: negative cross entropy)}]
                {\includegraphics[width=0.245\linewidth]{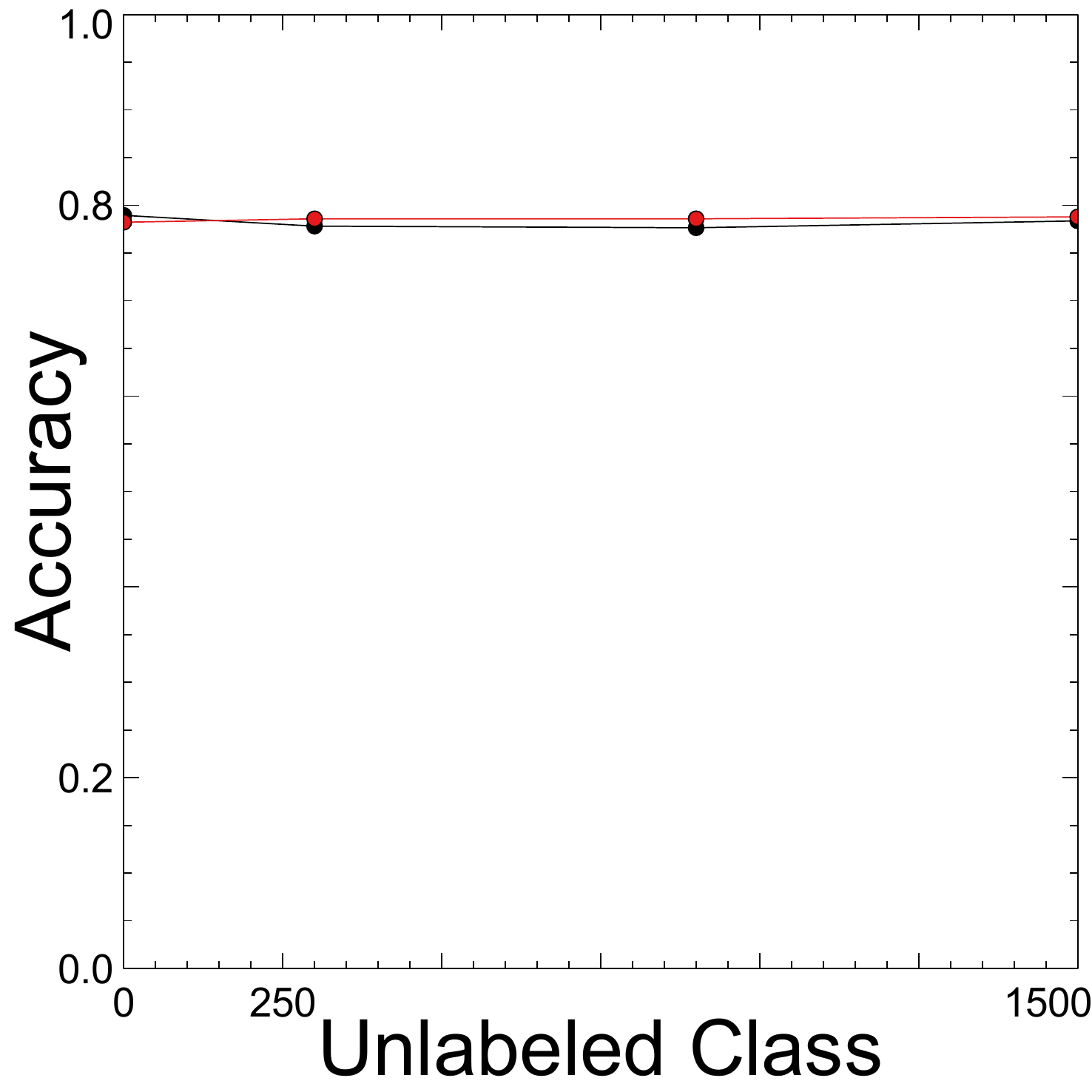}
                \includegraphics[width=0.245\linewidth]{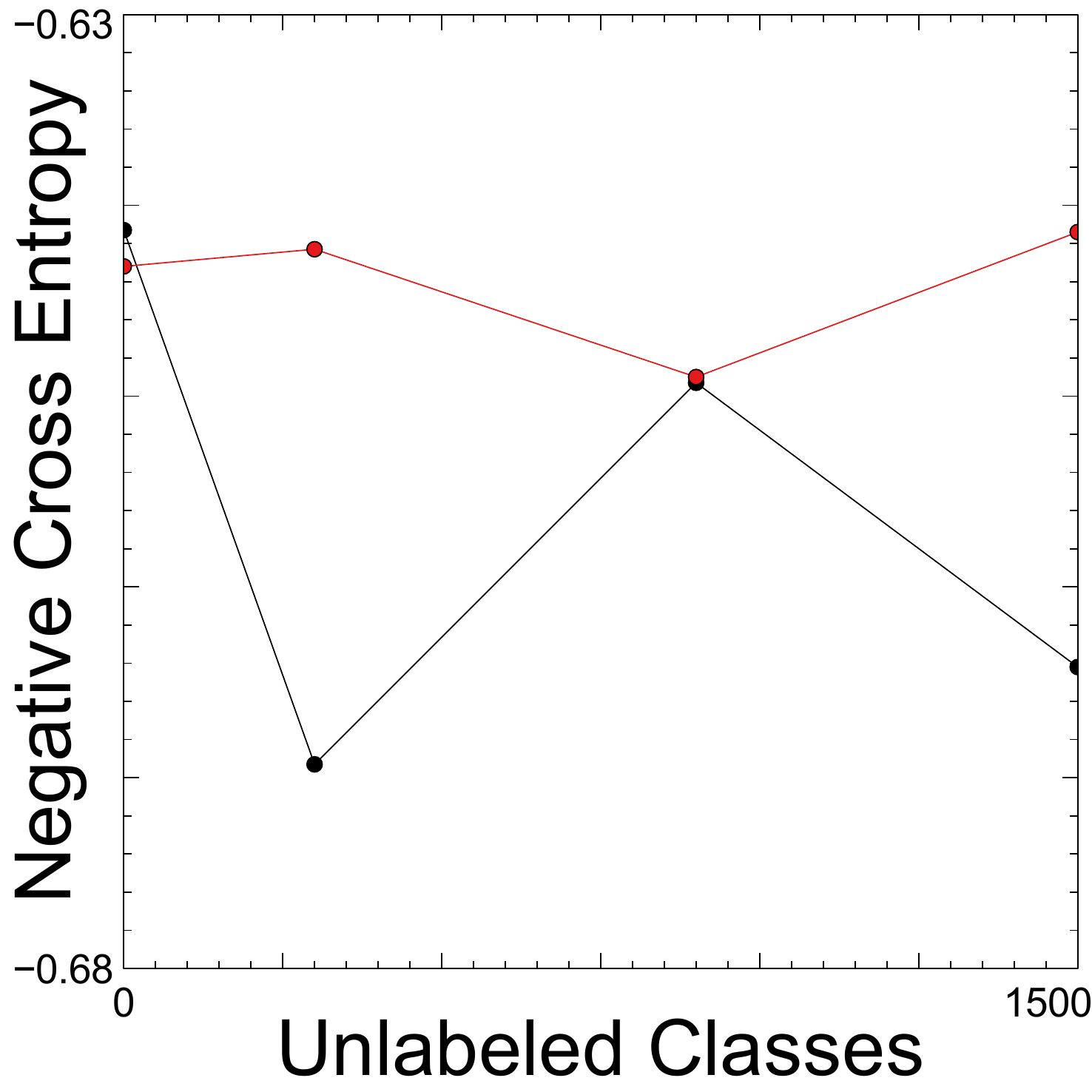}
                }
        \end{subfigure}%
          \vspace{-2mm}
        \caption{Predictive performance with increasing unlabeled classes when $\mathcal{C}_{\rm U}\cap \mathcal{C}_{\rm L} \neq \emptyset$}\label{fig:exp-adduc}
\end{figure*}

\par
\subsubsection{Results on mini-ImageNet}
Figure \ref{fig:exp-adduc} (a) shows the results on the mini-ImageNet dataset. 
The left endpoint represents the case when the classes of the unlabeled examples and those of the labeled examples are identical. Though this case satisfies the assumption of the baseline method~\cite{ren18}, i.e., $\mathcal{C}_{\rm U}=\mathcal{C}_{\rm L}$,
the proposed method exhibits slightly higher accuracy ($3.8\%$ higher) and negative cross entropy than the baseline method. These results show the superiority of our method even when the assumption of the baseline method \cite{ren18} holds.
In increasing number of the unlabeled classes which is not included in the labeled classes, the accuracy of our method slightly increases when more than $6$ classes are added, and the cross entropy steadily increases. These results again support our hypothesis, as in the previous section.
\par
\subsubsection{Results on VGG-Face}
Figure \ref{fig:exp-adduc} (b) shows the results on the VGG-Face dataset. We see that the accuracy and the negative cross entropy of the two methods are similar.
Note that at the left end, when all unlabeled examples belong to the labeled training classes, i.e., $\mathcal{C}_{\rm U}=\mathcal{C}_{\rm L}$, the baseline method slightly outperforms our method.
These results are not surprising because the case satisfies the assumption of the self-training of the baseline method
\cite{ren18}.
However, by adding examples of which underlying classes are not included in the labeled class $\mathcal{C}_{\rm L}$ to unlabeled training examples, the performance of the baseline method in both evaluation measures degrades, whereas that of our method improves.
These results again support our hypothesis as in the mini-ImageNet dataset and the experiments in the previous section. 

\subsection{Influence of Weight $\beta$ of the Marginal Likelihood}\label{sec:exp-beta}
We investigate the influence of weight $\beta$ of the marginal likelihood. 
Theoretically, the discriminative model delivers lower generalization loss with an infinite number of labeled examples, whereas a smaller number of the labeled examples reinforces the tendency that the generative model achieves lower generalization loss \cite{ng02}.
We confirm the advantage of our hybrid model, which takes an intermediate model of a pure discriminative model ($\beta=0$) and a pure generative model ($\beta=1$).
Note that the $\beta = 0$ corresponds to pure Prototypical Networks (implemented with Normalizing Flow), and the performance gain from $\beta = 0$ shows the contribution of the marginal density.

\par

For this analysis, we used the mini-ImageNet dataset. For the labeled examples and classes $\mathcal{C}_{\rm L}$, we used the common setting described in Section~\ref{sec:exp-data}.
For the unlabeled training set, we used all remaining examples, i.e., 60\% examples in a labeled class and all examples in an unlabeled class.


\begin{figure}[tbp]
\centering
                \includegraphics[width=0.48\linewidth]{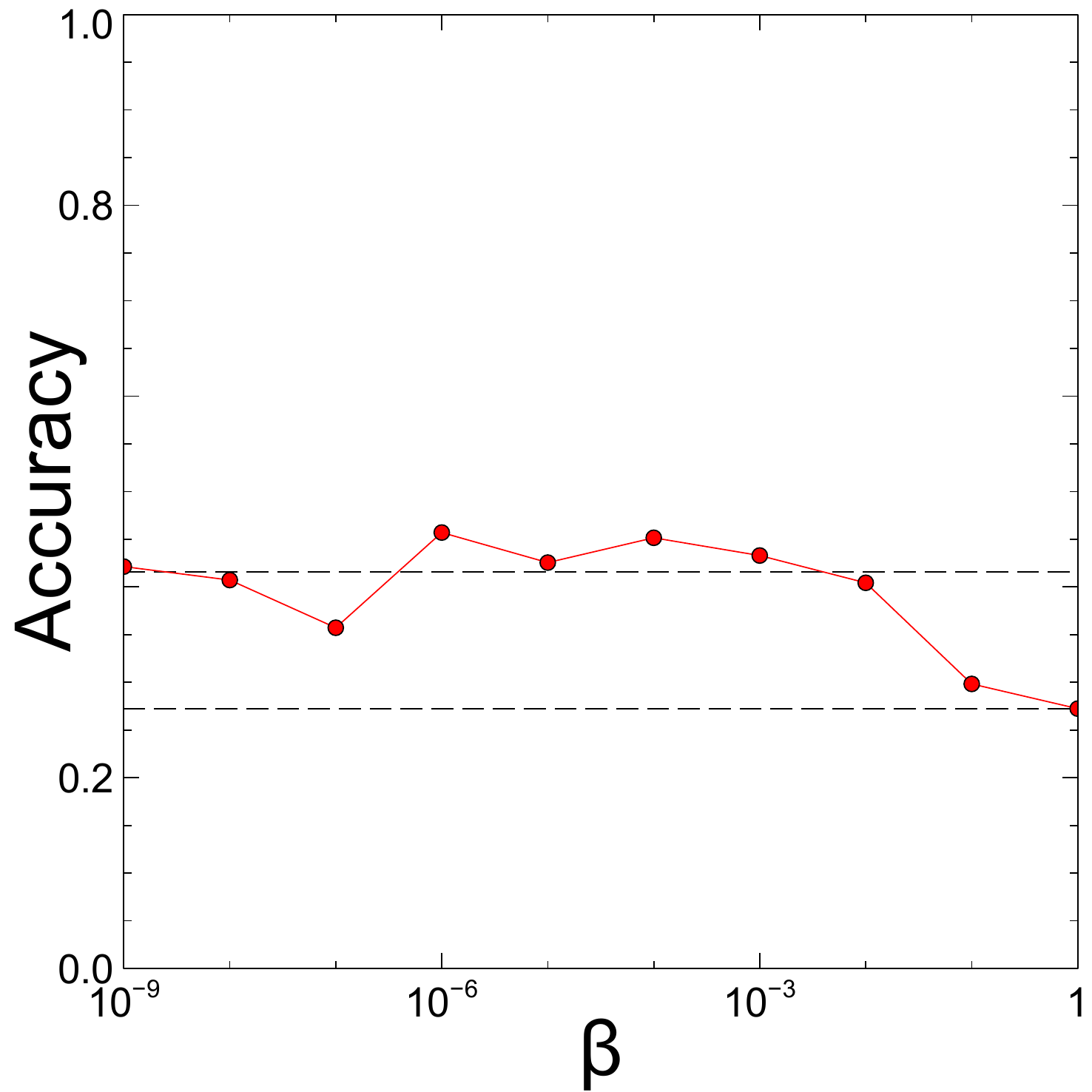}
                \includegraphics[width=0.48\linewidth]{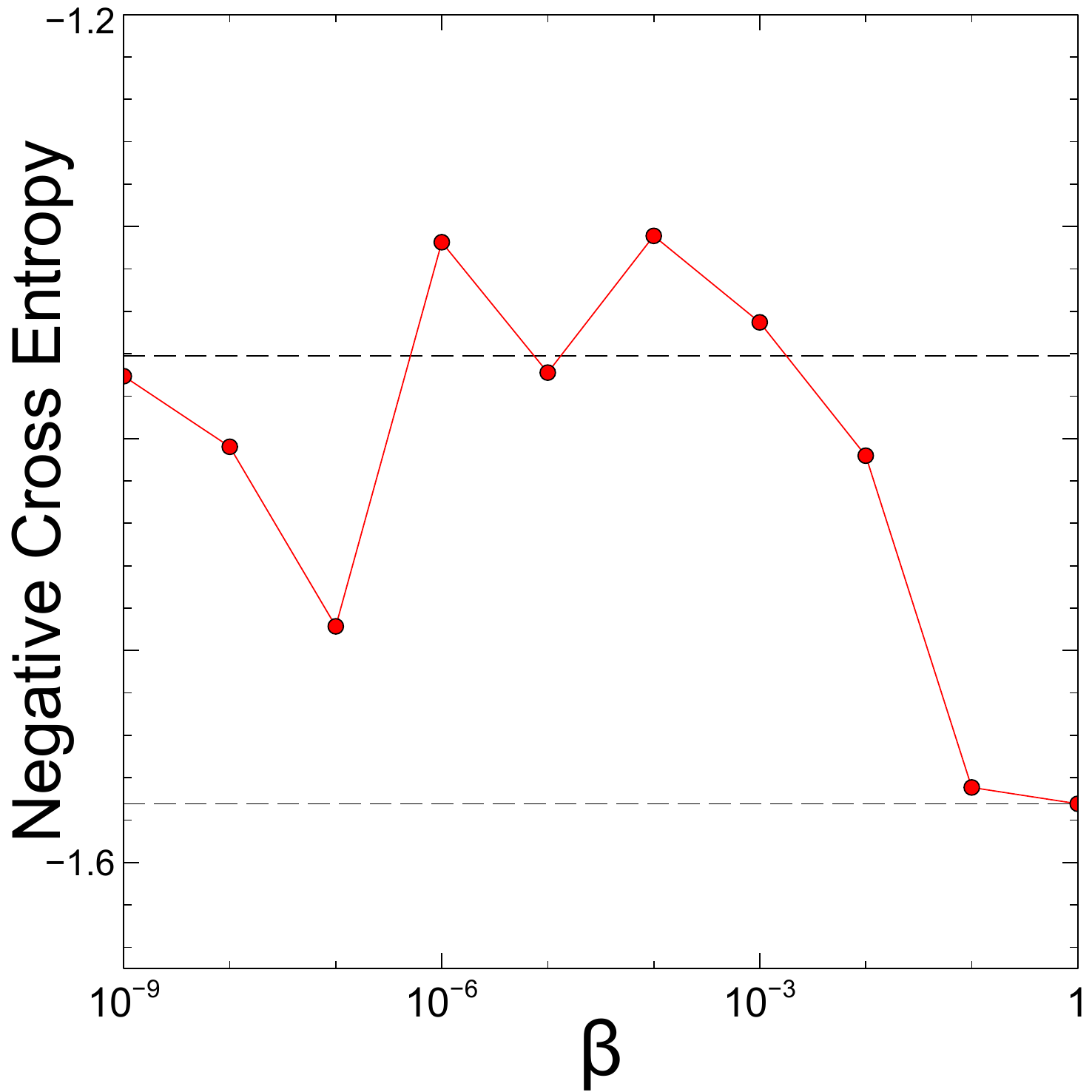}
        \caption{Predictive performance in terms of weight $\beta$ on marginal likelihood $p(\mathvc{x}|\mathvc{\theta})$. The cases of a pure discriminative model ($\beta=0$, the upper dashed line) and that of a pure generative model ($\beta=1$, the lower dashed line) are shown. Note that the intermediate models between the dashed lines show better accuracies.}\label{fig:result-beta}
  \end{figure}
Figure \ref{fig:result-beta} left shows the predictive accuracy in the test set
in terms of weight $\beta$ on marginal likelihood $p(\mathvc{x}|\mathvc{\theta})$.
The cases of a pure discriminative model ($\beta=0$) and that of a
pure generative model ($\beta=1$) are shown with dashed lines. 
We see that the intermediate models between the dashed lines show better accuracies and the negative cross entropies. 
These results confirm the analysis by Liang et al. that the best model that exhibits the lowest generalization loss exists between a discriminative model and a generative model unless there is an infinite number of training examples or the data is generated from marginal density $p(\mathvc{x}|\mathvc{\theta})$ \cite{liang08}.


\section{Conclusions}\label{sec:discussion}

We have proposed a new method which combines Prototypical Networks and Normalizing Flow in the framework of composite likelihood for the semi-supervised few-shot classification problem.
Experiments showed that by adding more unlabeled examples, the proposed method improves predictive distributions of the test classes even when there is no class overlap between the unlabeled and the labeled training examples. 
These facts show the effectiveness of our modeling which does not assume the classes of the unlabeled training examples are the same with the classes of the labeled examples.
Moreover, our experiments also show that the semi-supervised classification model also satisfies the conjecture that a hybrid model of discriminative and generative models exhibits better predictive performance when the labeled training examples are scarce~\cite{ng02, liang08}.
\bibliography{reference} 
\bibliographystyle{plain} 

\end{document}